\documentclass[lettersize,journal]{IEEEtran}
\usepackage{amsmath,amsfonts}
\usepackage{algorithmic}
\usepackage{algorithm}
\usepackage{array}
\usepackage[caption=false,font=normalsize,labelfont=sf,textfont=sf]{subfig}
\usepackage{textcomp}
\usepackage{stfloats}
\usepackage{url}
\usepackage{verbatim}
\usepackage{graphicx}
\usepackage{cite}
\usepackage{multirow}
\usepackage{graphicx}
\usepackage{tabularx}
\usepackage{booktabs}
\usepackage{xcolor}
\usepackage{url}
\usepackage{svg}
\begin{document}

\title{OceanMAE: A Foundation Model for Ocean Remote
Sensing}

\author{Viola-Joanna Stamer, Panagiotis~Agrafiotis,  Behnood Rasti,~\IEEEmembership{Senior Member, IEEE} and Beg{\"u}m~Demir,~\IEEEmembership{Senior Member, IEEE}
\thanks{The authors are with the Faculty of Electrical Engineering and Computer Science, Technische Universit{\"a}t Berlin, Germany and with the Berlin Institute for the Foundations of Learning and Data (BIFOLD), 10623 Berlin, Germany. The contribution of Panagiotis Agrafiotis is part of the MagicBathy project, funded by the European Union’s Horizon Europe research and innovation programme under the Marie Skłodowska-Curie Actions (Grant Agreement No. 101063294). \textit{(Co-corresponding authors: Behnood Rasti and Panagiotis Agrafiotis)}}
}

\markboth{Journal of \LaTeX\ Class Files,~Vol.~14, No.~8, August~2021}%
{Shell \MakeLowercase{\textit{et al.}}: A Sample Article Using IEEEtran.cls for IEEE Journals}


\maketitle

\begin{abstract}
Accurate ocean mapping is essential for applications such as bathymetry estimation, seabed characterization, marine litter detection, and ecosystem monitoring. However, ocean remote sensing (RS) remains constrained by limited labeled data and by the reduced transferability of models pre-trained mainly on land-dominated Earth observation imagery. In this paper, we propose OceanMAE, an ocean-specific masked autoencoder that extends standard MAE pre-training by integrating multispectral Sentinel-2 observations with physically meaningful ocean descriptors during self-supervised learning. By incorporating these auxiliary ocean features, OceanMAE is designed to learn more informative and ocean-aware latent representations from large-scale unlabeled data. To transfer these representations to downstream applications, we further employ a modified UNet-based framework for marine segmentation and bathymetry estimation. Pre-trained on the Hydro dataset \cite{Corley:2024}, OceanMAE is evaluated on MADOS \cite{mados} and MARIDA \cite{MARIDA} for marine pollutant and debris segmentation, and on MagicBathyNet \cite{agrafiotis2024magicbathynet} for bathymetry regression. The experiments show that OceanMAE yields the strongest gains on marine segmentation, while bathymetry benefits are competitive and task-dependent. In addition, an ablation against a standard MAE on MARIDA indicates that incorporating auxiliary ocean descriptors during pre-training improves downstream segmentation quality. These findings highlight the value of physically informed and domain-aligned self-supervised pre-training for ocean RS. Code and weights are publicly available at \url{https://git.tu-berlin.de/joanna.stamer/SSLORS2}.
\end{abstract}


\section{Introduction}
\IEEEPARstart{T}{he} ocean covers over 70\% of the Earth’s surface, produces nearly 50\% of global oxygen, and acts as a major carbon sink. Accurate monitoring of marine environments is therefore essential for detecting environmental hazards and tracking climate change. Although in situ measurements provide high accuracy, they are point-based, costly, and labor-intensive. Ocean remote sensing (RS) from airborne and space-borne platforms enables large-scale observations at lower cost, supports a wide range of spatial resolutions \cite{mandlburger2022review}, and generates terabytes of data daily \cite{wang2023deepblue}. Deep learning (DL) methods have shown strong potential for ocean-related tasks \cite{wang2023deepblue, chen2024satellite, khurram2025developments, agrafiotis2026seabed, hao2025deep}, but their performance is often limited by the scarcity of labeled marine data. Existing pre-trained models also struggle to capture dynamic water-surface phenomena, such as waves, sun glint, ripples, and the daily or seasonal variability of the sea \cite{manas2021seasonal, agrafiotis2024magicbathynet}. Moreover, annotating ocean imagery is time-consuming and expensive. Although many remote sensing foundation models (FMs) have recently emerged \cite{hong2024spectralgpt}, their transferability to ocean remote sensing remains limited, as most are pre-trained on data dominated by terrestrial scenes and are therefore not explicitly designed to capture the distinctive spectral and physical characteristics of water bodies.

Self-supervised learning (SSL) offers a promising alternative by leveraging large volumes of unlabeled data. Motivated by this, we introduce OceanMAE, an ocean-focused masked autoencoder that jointly exploits multispectral observations and physical oceanic features during pre-training. To transfer these representations to downstream applications, we further propose a modified UNet \cite{ronneberger2015u} for segmentation and a modified Bathy-UNet \cite{agrafiotis2024magicbathynet} for bathymetry regression, both equipped with a parallel embedding stream to integrate the learned features effectively. Finally, we evaluate different embedding initialization strategies across three representative ocean RS tasks: bathymetry regression, pollutants and sea-surface segmentation, and marine debris segmentation.

\section{Datasets}
\label{sec:datasets}

\subsection{Pre-Training Dataset}
To support downstream ocean RS applications with domain-specific pre-training data, we use the Hydro dataset \cite{Corley:2024}. Hydro comprises 100,000 sampled $256 \times 256$ Sentinel-2 L2A patches covering oceanic and coastal regions worldwide, sourced from the Planetary Computer STAC collection.

\subsection{Fine-Tuning Datasets} 
For fine-tuning, we use MARIDA \cite{MARIDA} for marine debris and water-related semantic segmentation, MADOS \cite{mados} for marine pollutant segmentation, and MagicBathyNet \cite{agrafiotis2024magicbathynet} for bathymetry estimation.

\subsubsection{MARIDA}
MARIDA is a benchmark dataset based on multispectral Sentinel-2 imagery for detecting and monitoring floating marine debris. It provides globally distributed annotations of real-world debris events together with co-occurring sea-surface features. The dataset contains more than 1,381 image patches and 837,377 annotated pixels derived from approximately 63 Sentinel-2 scenes acquired between 2015 and 2021. MARIDA defines 15 marine classes, and most patches are dominated by water.

\subsubsection{MADOS}
MADOS is designed to address two major marine pollutants: marine debris and oil spills. It is based on 174 Sentinel-2 L2A scenes acquired worldwide between 2015 and 2022 and contains approximately 1.5 million annotated pixels. The dataset includes 15 classes, covering oil spills and other water-related categories.

\subsubsection{MagicBathyNet}
For bathymetry regression, we use the public multimodal MagicBathyNet benchmark for shallow-water depth estimation. The dataset covers two contrasting coastal environments: Agia Napa (Mediterranean Sea, maximum depth $\approx -30.3\,\text{m}$, 35 images) and Puck Lagoon (Baltic Sea, maximum depth $\approx -10.6\,\text{m}$, 2822 images), which differ in water properties and seabed composition.

\section{Methodology}
\label{sec:methodology}
The proposed framework follows a two-stage training paradigm. In the first stage, OceanMAE is pre-trained in a self-supervised manner on large-scale unlabeled ocean imagery and physical oceanic features. In the second stage, the learned encoder is transferred to downstream models through specialized UNet-based architectures for segmentation and bathymetry estimation. Let the input image be \(X \in \mathbb{R}^{H \times W \times C}\), where \(H\) and \(W\) denote the spatial dimensions and \(C=11\) is the number of spectral bands. The downstream target is denoted by \(Y\), representing either a pixel-wise label map for segmentation or a depth map for bathymetry estimation.

\subsection{Pre-Training Phase: Proposed OceanMAE}
The pre-training stage follows the generative self-supervised paradigm of MAEs to learn robust feature representations \(z\).

\subsubsection{A Brief Recall of MAE \cite{MAE}}
An input image \(X\) is tokenized into \(n\) patches, \(X=\{p_1,\dots,p_n\}\). A subset of patches, \(X_{\text{mask}}\), is randomly masked using a masking ratio of \(90\%\), leaving the visible patches \(X_{\text{vis}} = X \setminus X_{\text{mask}}\). The MAE encoder \(E\) processes only the visible patches together with learned positional embeddings \(\mathcal{L}_{\text{pos}}\) to produce a latent representation \(z \in \mathbb{R}^{D_{\text{embed}}}\):
\begin{equation}
z = E(X_{\text{vis}} + \mathcal{L}_{\text{pos}}).
\end{equation}
The MAE decoder \(D\) then reconstructs the full image \(\hat{X}\) from the latent representation \(z\) and the mask tokens \(\mathcal{M}\).

\subsubsection{First Stage: Methodological Overview of OceanMAE}
OceanMAE adapts the MAE framework to ocean imagery by augmenting representation learning with external oceanic variables. Its objective is to learn informative latent features not only from masked image reconstruction but also from auxiliary physical descriptors.

\begin{figure}
    \centering
    \includegraphics[scale=0.2]{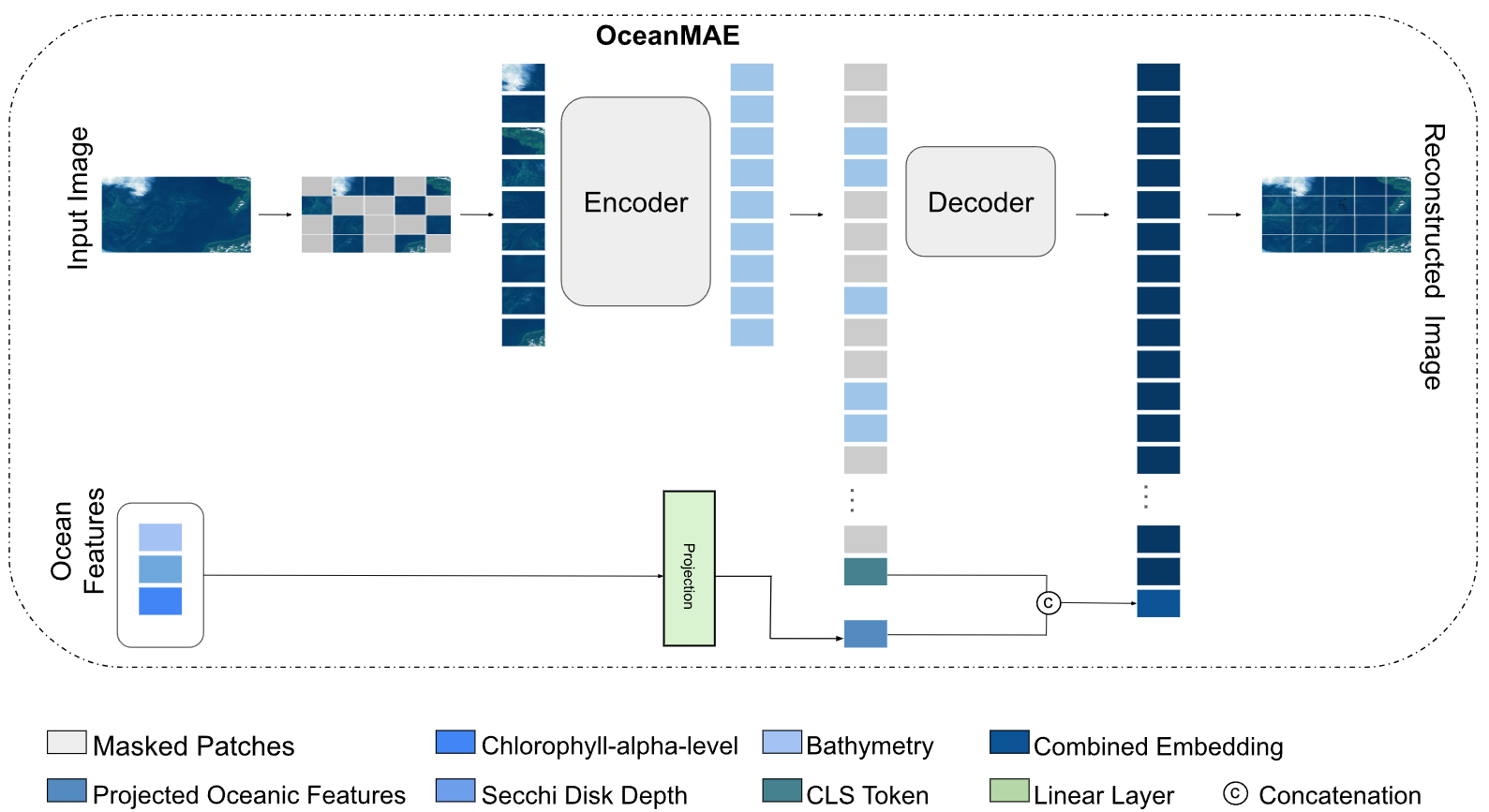}
    \caption{Overview of the OceanMAE architecture adapted from \cite{MAE}. Input patches are masked with a ratio of \(90\%\). The visible patches are processed by the encoder and combined with learnable mask tokens to reconstruct the full image. In parallel, external ocean features are projected and concatenated with the \([\mathrm{CLS}]\) token to form a multimodal representation used during reconstruction.}
    \label{fig:ocean_mae}
\end{figure}

As illustrated in Fig.~\ref{fig:ocean_mae}, OceanMAE incorporates external oceanic features \(F_{\text{ocean}}\), such as bathymetry, chlorophyll level, and Secchi depth. These variables complement Sentinel-2 imagery by providing physically meaningful context. The encoder output \({E}_{\text{CLS}}\) is fused with the auxiliary features \(F_{\text{ocean}} \in \mathbb{R}^{B \times N_{\text{ocean}}}\) in two steps:
\begin{enumerate}
    \item \textit{Projection:} The raw ocean features are linearly projected to the visual embedding dimension \(D_{\text{embed}}\):
    \begin{equation}
    \hat{F}_{\text{ocean}} = F_{\text{ocean}}W_{\text{proj}} + b_{\text{proj}},
    \qquad
    \hat{F}_{\text{ocean}} \in \mathbb{R}^{B \times D_{\text{embed}}}.
    \end{equation}
    
    \item \textit{Concatenation:} The projected features are concatenated with the \({E}_{\text{CLS}}\) token along the feature dimension to form the multimodal embedding \(E_{\text{combined}}\):
    \begin{equation}
    E_{\text{combined}} = E_{\text{CLS}} \bigoplus \hat{F}_{\text{ocean}},
    \qquad
    E_{\text{combined}} \in \mathbb{R}^{B \times 2D_{\text{embed}}}.
    \end{equation}
\end{enumerate}

This fusion allows the decoder to exploit both learned visual patterns and physical oceanic context during reconstruction. For example, bathymetric information can help distinguish deep-water from shallow-water regions even when their spectral appearance is similar. In the downstream stage, the visual token \(E_{\text{CLS}}\) is retained as the latent representation \(z\) used for transfer.

\subsubsection{Reconstruction Loss}
The training objective is the Mean Squared Error (MSE) between the original masked patches \(x_i\) and their reconstructions \(\hat{x}_i\), computed over the set of masked patch indices \(\mathcal{M}\):
\begin{equation}
L_{\text{MSE}} = \frac{1}{|\mathcal{M}|}\sum_{i \in \mathcal{M}} (x_i - \hat{x}_i)^2.
\end{equation}
Here, \(|\mathcal{M}|\) denotes the number of masked patches. Minimizing this loss forces the encoder \(E\) to learn informative representations from incomplete observations. The pre-trained encoder is then transferred to the fine-tuning stage.

\subsection{Fine-Tuning Stage: Adapted UNet and Bathy-UNet Architectures}
The downstream models \(U\) are designed to exploit the pre-trained embedding \(z\). They are trained on MADOS, MARIDA, and MagicBathyNet for pollutant segmentation, marine debris segmentation, and bathymetry regression, respectively, to predict the output map \(\hat{Y}\).

\subsubsection{Embedding Initialization Strategies}
The pre-trained encoder \(E\) is used to generate the embedding \(z\) from each downstream input image. We investigate three initialization strategies:
\begin{enumerate}
    \item \textit{Random:} \(E\) is randomly initialized, the resulting embeddings \(z\) are computed once, and then kept fixed \((\eta = 0)\).
    \item \textit{Frozen Embedding (FE):} \(E\) is initialized with pre-trained weights and kept fixed during downstream training \((\eta = 0)\).
    \item \textit{Fully Finetuned (FF):} \(E\) is initialized with pre-trained weights and updated during downstream training \((\eta > 0)\).
\end{enumerate}

\subsubsection{Second Stage: UNet and Bathy-UNet with Parallel Embedding Stream}
The adapted UNet and Bathy-UNet, denoted by \(U\), consist of a convolutional encoder \(E_{\text{conv}}\), a convolutional decoder \(D_{\text{conv}}\), and a parallel embedding projection path \(P_{\text{embed}}\). The final prediction \(\hat{Y}\) is given by
\begin{equation}
\hat{Y} = U(X,z) = D_{\text{conv}}\!\left(\mathcal{F}\!\left(E_{\text{conv}}(X), P_{\text{embed}}(z)\right), S\right),
\end{equation}
where \(S\) denotes the set of skip connections and \(\mathcal{F}\) is the feature-fusion operator.

\paragraph{Contracting Path (Encoder)}
The standard UNet encoder \(E_{\text{conv}}: \mathbb{R}^{H \times W \times C} \rightarrow \mathbb{R}^{H_b \times W_b \times C_b}\) extracts hierarchical convolutional features \(F_{\text{conv}}\). At each stage \(k\), it applies two \(3 \times 3\) convolutions with ReLU activations, followed by \(2 \times 2\) max pooling:
\begin{equation}
\tilde{F}^{(k)}=\mathrm{ConvReLU}\!\big(\mathrm{ConvReLU}(F^{(k-1)})\big)
\end{equation}
\begin{equation}
F_{\text{conv}}^{(k)}=\mathrm{MaxPool}(\tilde{F}^{(k)})
\end{equation}

\paragraph{Parallel Embedding Path}
The embedding path \(P_{\text{embed}}: \mathbb{R}^{D_{\text{embed}}} \rightarrow \mathbb{R}^{H_b \times W_b \times C_b'}\) maps the latent vector \(z\) to a spatial feature map \(F_{\text{embed}}\) compatible with the UNet bottleneck. This path consists of three steps:
\begin{enumerate}
    \item A linear projection maps \(z\) from \(D_{\text{embed}}\) dimensions to \(H_b \cdot W_b \cdot C_{\text{init}}\).
    \item The projected vector is reshaped into a spatial tensor \(\tilde{F}_{\text{embed}} \in \mathbb{R}^{H_b \times W_b \times C_{\text{init}}}\).
    \item A sequence of convolutional layers refines this tensor to produce \(F_{\text{embed}} \in \mathbb{R}^{H_b \times W_b \times C_b'}\).
\end{enumerate}
\begin{equation}
F_{\text{embed}} = P_{\text{embed}}(z).
\end{equation}

\begin{figure}[htbp]
  \centering
  \includegraphics[scale=0.28]{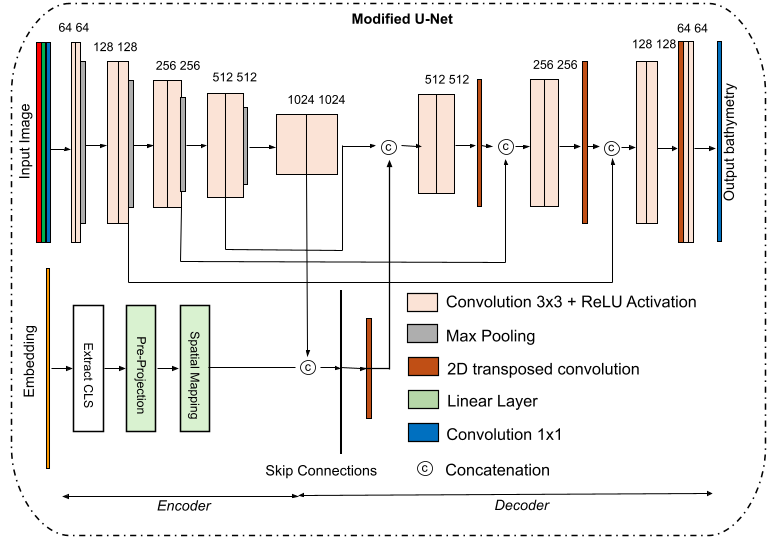}
  \caption{Architecture of the modified UNet for downstream ocean tasks. The model follows an encoder-decoder design with skip connections to preserve spatial detail. The OceanMAE embedding is projected and fused at the bottleneck, enabling the downstream network to exploit pre-trained ocean-aware representations for both dense and sparse prediction tasks.}
  \label{fig:unet}
\end{figure}

\paragraph{Bottleneck Feature Fusion}
At the bottleneck, the convolutional features \(F_{\text{conv}}\) and the projected embedding features \(F_{\text{embed}}\) are fused by \(\mathcal{F}\):
\begin{equation}
F_{\text{fused}} = \mathcal{F}(F_{\text{conv}}, F_{\text{embed}}).
\end{equation}
If \(\mathcal{F}\) is concatenation, then
\begin{equation}
F_{\text{fused}} \in \mathbb{R}^{H_b \times W_b \times (C_b + C_b')}.
\end{equation}
This fusion enables the decoder to combine local spatial detail from \(E_{\text{conv}}\) with global context from the pre-trained OceanMAE embedding, as illustrated in Fig.~\ref{fig:unet}.

\section{Experiments}
\label{sec:experiments}

\subsection{Experimental Design}
During fine-tuning, different embedding initialization strategies were evaluated. A ViT-Base architecture \cite{dosovitskiy2021imageworth16x16words} with embedding dimension $D=768$ was used as the pre-trained backbone to extract the latent representations. To ensure consistency, images from the downstream datasets were transformed and normalized in the same way as during pre-training before being passed through the encoder. The resulting embeddings were linearly projected and reshaped to match the bottleneck resolution of UNet and Bathy-UNet, fused with the deepest feature map, and then mapped to the required channel dimension before decoding. 

The proposed models were evaluated on three downstream tasks: pollutant and sea-surface segmentation on MADOS, marine debris segmentation on MARIDA, and bathymetry regression on MagicBathyNet. OceanMAE was compared against established Earth observation self-supervised models, including FGMAE \cite{fgmae}, SatMAE \cite{cong2022satmae}, and SSL4EO \cite{wang2023ssl4eos12largescalemultimodalmultitemporal}. For segmentation, we report Pixel Accuracy (PA), mean Intersection over Union (mIoU), and Macro F1. For bathymetry regression, we report Mean Absolute Error (MAE), Root Mean Square Error (RMSE), and standard deviation (StdDev).

\subsection{Configuration Details}
In the pre-training stage, OceanMAE was trained on the Hydro dataset using an NVIDIA A100-SXM4-80GB GPU. The input size was $224 \times 224$ with a patch size of $16 \times 16$. In addition, a patch size of $4 \times 4$ was evaluated for the downstream task of pollutant and sea-surface segmentation. A masking ratio of $90\%$ was applied. The decoder had dimension $512$ and was optimized using an MSE reconstruction loss. Training was performed with AdamW and a multi-step learning-rate schedule, reducing the learning rate by a factor of $0.1$ to $1 \times 10^{-4}$ at epochs $5$, $10$, and $15$. The model was trained for $100$ epochs with a batch size of $64$. Different learning rates, schedules, and epoch numbers were tested, and this configuration yielded the best overall performance. The total pre-training time of OceanMAE was approximately $15$--$17.5$ hours.

\subsection{Bathymetry Regression on MagicBathyNet}
\label{sec:results:bathymetry}
This experiment evaluates the effect of embedding initialization on bathymetry regression in two contrasting regions, Agia Napa ($35$ images) and Puck Lagoon ($2822$ images). The results are reported in Table \ref{tab:bathymetry_results}. Pre-trained embeddings generally improve performance over random initialization and yield more stable predictions than the baseline in most settings. On the small Agia Napa subset, the Fully Finetuned strategy achieved the best OceanMAE performance, indicating that pre-training is particularly helpful when labeled data are scarce. On the larger Puck Lagoon subset, the UNet-bathy baseline achieved the lowest MAE, whereas OceanMAE + FF obtained the best RMSE and StdDev, suggesting a trade-off between absolute error and prediction stability. Overall, the impact of OceanMAE on bathymetry is competitive but region- and dataset size-dependent, likely reflecting differences in water properties and seabed characteristics.

\begin{table}
\centering
\caption{Bathymetry regression results on MagicBathyNet. Lower is better.}
\label{tab:bathymetry_results}
\begin{tabular}{llccc}
\toprule
Region & Model & MAE & RMSE & StdDev \\ 
\midrule
Agia Napa & UNet-bathy \cite{agrafiotis2024magicbathynet} & 0.694 & 1.068 & 0.940 \\ 
          & OceanMAE + Random & 0.560 & 0.731 & 0.613 \\ 
          & OceanMAE + FE & 0.536 & 0.714 & 0.594 \\ 
          & SatMAE + FF & 0.623 & 0.812 & 0.721 \\ 
          & FGMAE + FF & 0.651 & 0.868 & 0.712 \\
          & SSL4EO + FF & 0.612 & 0.823 & 0.722 \\ 
          & OceanMAE + FF & \textbf{0.479} & \textbf{0.637} & \textbf{0.591} \\ 
\midrule
Puck Lagoon & UNet-bathy \cite{agrafiotis2024magicbathynet} & \textbf{0.493} & 0.907 & 0.874 \\ 
            & OceanMAE + Random & 0.874 & 1.249 & 0.885 \\ 
            & OceanMAE + FE & 0.689 & 0.879 & 0.675 \\ 
            & SatMAE + FF & 0.701 & 1.104 & 0.921 \\ 
            & FGMAE + FF & 0.682 & 0.941 & 0.810 \\
            & SSL4EO + FF & 0.783 & 1.201 & 1.001 \\ 
            & OceanMAE + FF & 0.555 & \textbf{0.757} & \textbf{0.536} \\ 
\bottomrule
\end{tabular}
\end{table}

\subsection{Marine Debris Segmentation on MARIDA}
\label{sec:results:debris}
For marine debris segmentation, we first compare different OceanMAE initialization strategies against external EO self-supervised models. The results are summarized in Table \ref{tab:debris_results}. Among the OceanMAE variants, the Fully Finetuned strategy achieved the best PA and Macro F1, while the Frozen Embedding strategy yielded slightly lower but still competitive performance. Compared with the original UNet baseline \cite{MARIDA}, OceanMAE + FF improves PA from $0.69$ to $0.77$ and Macro F1 from $0.69$ to $0.72$. In contrast, external models such as FGMAE achieve higher mIoU, indicating that the relative benefit depends on the evaluation metric. Overall, OceanMAE provides the strongest gains on overall pixel classification and class-balanced performance on MARIDA.

\begin{table}[h!]
    \centering
    \caption{Segmentation results on MARIDA. Higher is better.}
    \label{tab:debris_results}
\begin{tabular}{lccc}
    \toprule
    Model & PA & mIoU & Macro F1 \\ 
    \midrule
    UNet \cite{MARIDA} & 0.69 & 0.57 & 0.69 \\
    OceanMAE + Random & 0.69 & 0.57 & 0.68 \\ 
    OceanMAE + FE & 0.72 & 0.58 & 0.69 \\
    SatMAE + FF & 0.54 & 0.67 & 0.64 \\
    FGMAE + FF & 0.55 & \textbf{0.68} & 0.66 \\
    SSL4EO + FF & 0.51 & 0.65 & 0.64 \\
    OceanMAE + FF & \textbf{0.77} & 0.61 & \textbf{0.72} \\
    \bottomrule
\end{tabular}
\end{table}

\subsection{Pollutants and Sea-Surface Segmentation on MADOS}
\label{sec:results:oilspill}
For pollutants and sea-surface segmentation, we analyze the effect of initialization strategy and patch size, and then compare the best-performing configuration with state-of-the-art methods.

\subsubsection{Sensitivity Analysis: Initialization Strategy and Patch Size}
To assess the robustness of the proposed model, we evaluate two initialization strategies, Frozen Embedding and Fully Finetuned, with patch sizes of $16 \times 16$ and $4 \times 4$, respectively. The results are reported in Table \ref{tab:oil_spill_sensitivity}.

\begin{table}[h!]
    \centering
    \caption{Sensitivity analysis on MADOS. Higher is better.}
    \label{tab:oil_spill_sensitivity}
    \begin{tabular}{lccc}
        \toprule
        Model & PA & mIoU & Macro F1 \\
        \midrule
        OceanMAE + FE (Patch-16) & 87.4 & \textbf{70.8} & \textbf{81.3} \\
        OceanMAE + FF (Patch-16) & 86.5 & 67.6 & 79.3 \\
        \midrule
        OceanMAE + FE (Patch-4) & \textbf{88.7} & 62.4 & 73.8 \\
        OceanMAE + FF (Patch-4) & 87.2 & 63.4 & 75.4 \\
        \bottomrule
    \end{tabular}
\end{table}

Two trends can be observed. First, freezing the embeddings leads to better segmentation quality than full fine-tuning, especially for Patch-16. In particular, OceanMAE + FE (Patch-16) improves over OceanMAE + FF (Patch-16) by $3.2$ points in mIoU and $2.0$ points in Macro F1. Second, Patch-16 yields consistently better mIoU and Macro F1 than Patch-4, indicating that the coarser patching strategy is better suited to this task.

\subsubsection{Comparison with State-of-the-Art}
The best-performing configuration, OceanMAE + FE (Patch-16), is compared against the baselines and state-of-the-art methods from \cite{mados}. The results are summarized in Table \ref{tab:oil_spill_sota}.

\begin{table}[h!]
    \centering
    \caption{Comparison with state-of-the-art on MADOS. Higher is better.}
    \label{tab:oil_spill_sota}
    \begin{tabular}{lccc}
        \toprule
        Model & PA & mIoU & Macro F1 \\
        \midrule
        Random Forest \cite{mados} & 67.1 & 43.9 & 56.6 \\
        UNet \cite{mados} & 82.9 & 51.0 & 63.8 \\
        SatMAE & 85.3 & 60.0 & 71.2 \\
        FGMAE & 82.6 & 65.1 & 77.1 \\
        SSL4EO & 83.0 & 63.6 & 74.8 \\
        MariNeXt (SOTA) \cite{mados} & \textbf{89.1} & 64.3 & 76.0 \\
        OceanMAE + FE (Patch-16) & 87.4 & \textbf{70.8} & \textbf{81.3} \\
        \bottomrule
    \end{tabular}
\end{table}

\begin{figure}[htbp]
  \centering
  \includegraphics[scale=0.27]{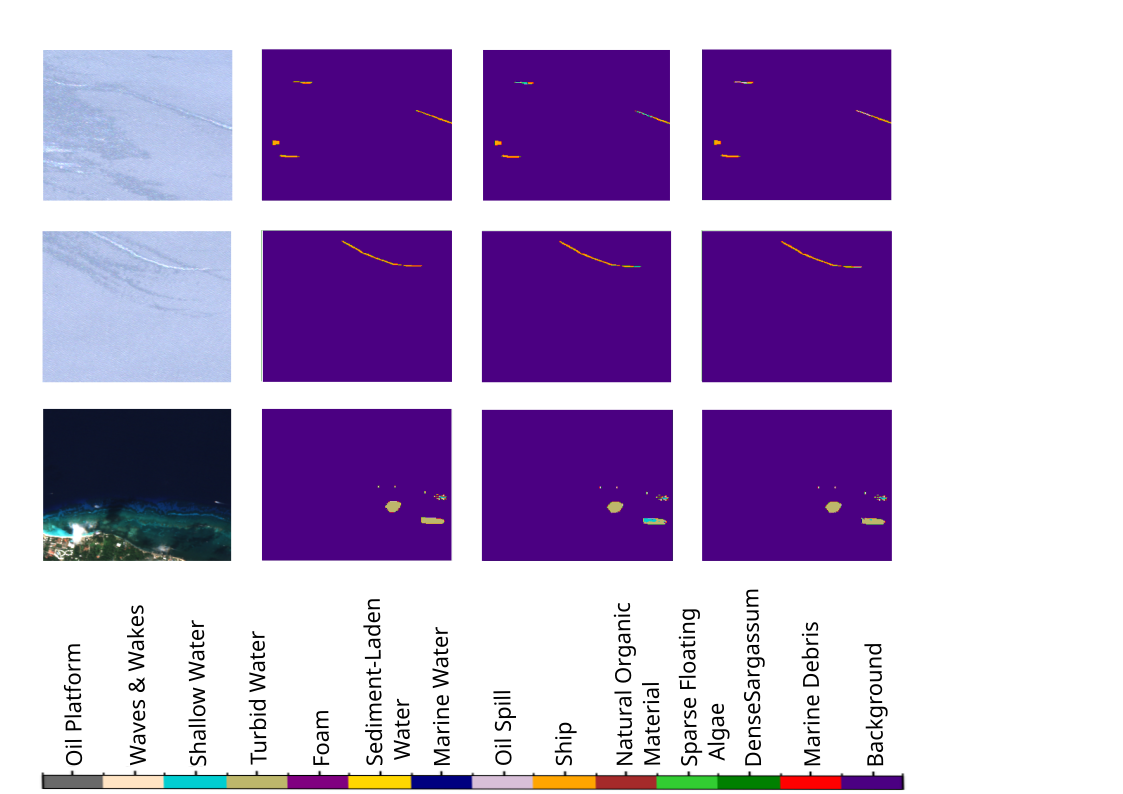}
  \caption{Qualitative comparison of pollutants and sea-surface segmentation results. From left to right: (a) RGB input, (b) ground-truth mask, (c) UNet baseline, and (d) the proposed Frozen Embedding strategy.}
  \label{fig:mados_qualitative}
\end{figure}

OceanMAE + FE (Patch-16) achieves the best mIoU and Macro F1, outperforming the previous best model, MariNeXt, by $6.5$ points in mIoU and $5.3$ points in Macro F1. Although MariNeXt attains the highest PA, OceanMAE provides the strongest overall segmentation quality on this benchmark. As shown in Fig.~\ref{fig:mados_qualitative}, the pre-trained embeddings produce cleaner predictions with reduced noise and sharper object boundaries than the UNet baseline.

\subsubsection{Ablation: Standard MAE vs. OceanMAE on MARIDA}
To isolate the effect of the auxiliary ocean descriptors used during pre-training, we compare OceanMAE + FF against a standard MAE + FF under the same downstream setting. The results are shown in Table \ref{tab:marida_ablation}. OceanMAE improves PA and Macro F1 over the standard MAE, indicating that physically informed pre-training benefits downstream marine segmentation. At the same time, the standard MAE attains higher mIoU, suggesting that the impact of the auxiliary descriptors is metric-dependent rather than uniform across all segmentation criteria.

\begin{table}[h!]
    \centering
    \caption{Ablation on MARIDA: effect of auxiliary ocean descriptors during pre-training. Higher is better.}
    \label{tab:marida_ablation}
\begin{tabular}{lccc}
    \toprule
    Model & PA & mIoU & Macro F1 \\
    \midrule
    MAE + FF & 0.48 & \textbf{0.68} & 0.59 \\
    OceanMAE + FF & \textbf{0.77} & 0.61 & \textbf{0.72} \\
    \bottomrule
\end{tabular}
\end{table}

The MARIDA ablation provides evidence that the auxiliary ocean descriptors used during pre-training affect the learned representation. Compared with a standard MAE, OceanMAE improves PA and Macro F1, which suggests that physically informed pre-training helps the model better separate marine debris from visually similar sea-surface classes. At the same time, the higher mIoU of the standard MAE indicates that the contribution of the auxiliary descriptors is not uniform across all metrics, but mainly reflected in overall classification quality and class-balanced performance.


\section{Conclusion}
\label{sec:conclusion}
In this work, we introduce a two stage approach. In the first stage we develop OceanMAE as a foundation model for ocean remote sensing and in the second stage we propose a modified UNet architecture to leverage the pre-trained representations. The results demonstrate that FM enhance performance and transferability across diverse oceanic tasks, including pollutants and sea surface features segmentation, marine debris segmentation, and bathymetry regression. Our experiments emphasize the importance of task- and domain-aligned SSL strategies. Moreover, the choice of initialization strategy, frozen or fully finetuned, should be guided by the specific task characteristics, differing notably between sparse-object segmentation and dense regression tasks, as well as by the availability of labeled training data. Overall, leveraging large-scale unlabeled oceanic data through SSL offers an effective solution to the label-scarcity challenge prevalent in marine environments.

\bibliographystyle{IEEEtran}
\bibliography{IEEEabrv,refs}

\vfill
\end{document}